\documentclass[runningheads]{llncs}
\usepackage[T1]{fontenc}
\usepackage{graphicx}
\usepackage{lmodern}
\usepackage{amssymb}
\usepackage{mathtools}
\usepackage{amsmath}
\usepackage{tikz}
\usepackage{pgf}
\usepackage{booktabs}

\usetikzlibrary{shapes,shadows,arrows,positioning,angles,quotes}
\tikzset{My Arrow Style/.style={single arrow, fill=black!15, anchor=base, align=center,text width=2.3cm}}
\tikzstyle{arrow} = [thick,->,>=stealth]
\usetikzlibrary{calc,trees,positioning,arrows,fit,shapes,calc}

\tikzstyle{startstop} = [rectangle, rounded corners, minimum width=1.5cm, minimum height=0.5cm,text centered, draw=black, fill=red!30]
\tikzstyle{io} = [trapezium, trapezium left angle=70, trapezium right angle=110, minimum width=0.5cm, minimum height=0.5cm, text centered, draw=black, fill=blue!30]

\tikzstyle{process} = [rectangle, minimum width=3cm, minimum height=0.5cm, text centered, draw=black, fill=orange!30]
\tikzstyle{decision} = [diamond, minimum width=0.5cm, minimum height=0.1cm, text centered, draw=black, fill=green!30]
\tikzstyle{process2} = [rectangle, minimum width=1cm, minimum height=0.5cm, text centered, draw=black, fill=orange!30]
\tikzstyle{arrow} = [thick,->,>=stealth]
\begin{document}
\title{Enhancing RL Safety with Counterfactual LLM Reasoning}
%
%
\author{Dennis Gross \and Helge Spieker
}
\authorrunning{Gross et al.}
%
\institute{Simula Research Laboratory
}
\maketitle              
\begin{abstract}
\emph{Reinforcement learning (RL)} policies may exhibit unsafe behavior and are hard to explain. We use counterfactual large language model reasoning to enhance RL policy safety post-training.
We show that our approach improves and helps to explain the RL policy safety.
\end{abstract}
\section{Introduction}
\emph{Reinforcement learning (RL)} has transformed technology~\cite{mnih2013playing}.

An RL agent learns a \emph{policy} to achieve an objective by acting and receiving rewards from the environment. A \emph{neural network (NN)} represents the policy, mapping state observations to actions, with each observation consisting of features characterizing the environment~\cite{DBLP:conf/setta/Gross22}.

Unfortunately, learned policies are not guaranteed to avoid \emph{unsafe behavior}~\cite{DBLP:journals/jmlr/GarciaF15}, as rewards often do not fully capture complex safety requirements~\cite{DBLP:journals/aamas/VamplewSKRRRHHM22}.

To resolve the issue mentioned above, formal verification methods like \emph{model checking} have been proposed to reason about the safety of RL~\cite{DBLP:conf/formats/HasanbeigKA20,DBLP:conf/tacas/HahnPSSTW19}.
Model checking is not limited by properties that can be expressed by rewards but support a broader range of properties that can be expressed by \emph{probabilistic computation tree logic (PCTL)} \cite{DBLP:journals/fac/HanssonJ94}.

However, trained NN policies obscure their inner workings, and it is essential in the context of explainable RL~\cite{DBLP:journals/ml/Bekkemoen24} to understand them.
Explaining RL policies through counterfactual action outcomes can help non-experts understand policy preferences by depicting the trade-offs between alternative actions~\cite{DBLP:conf/aaai/AmitaiSA24}.
A counterfactual explanation answers the question ``Why \emph{action 1} rather than \emph{action 2}?'', where \emph{action 1} is the fact that occurred and \emph{action 2} is a hypothetical alternative that the user might have expected~\cite{lipton1990contrastive}.

Unfortunately, it may be difficult to decide which state to look at for a non-expert and which alternative action to choose~\cite{DBLP:conf/aaai/AmitaiA22}.

Here, \emph{large language models (LLMs)} may help explain RL policy outcomes~\cite{DBLP:journals/corr/abs-2310-05797,lu2024mental}.
LLMs are designed to understand and generate human-like text by learning from vast text data.

In this work, we apply model checking to identify states that may lead to safety violations because of the policy action selection and use an LLM to explain and determine which alternative action may be a better choice.

Our approach takes three inputs: a \emph{Markov Decision Process (MDP)} representing the RL environment, a trained policy, and a PCTL formula for safety measurements.
We \emph{incrementally build} only the reachable parts of the MDP, guided by the trained policy~\cite{DBLP:conf/setta/Gross22}.
We verify the policy's safety using the \emph{Storm} model checker~\cite{DBLP:journals/sttt/HenselJKQV22} and the PCTL formula.
From the verified model, we extract states that directly lead to a safety violation through a policy action.

For each extracted state, we provide the RL environment information and the state with its action leading to a safety violation to the LLM, asking it to explain the mistake and suggest a safer alternative.

Then, we reverify the policy with the action alternatives concerning safety properties in the RL environment.

In experiments, we show that an LLM can indeed explain failures, propose valid alternative actions, and improve the safety performance of trained RL policies by forcing the policy to select the LLM alternative action.
We compare our approach to a baseline approach~\cite{DBLP:conf/aaai/AmitaiSA24} that always chooses naively the second most favored action at the current state.

To summarize our \textbf{main contribution}, we combine model checking with explainable RL and LLMs to explain safe policy decision-making and propose explainable alternative action selections at safety-critical states that improve the safety performance of the trained RL policies post-training.

\paragraph{Related work.}
Although there is work combining LLMs with RL~\cite{DBLP:conf/icml/DuWWCDA0A23}, work that focuses on counterfactual reasoning~\cite{DBLP:conf/aaai/AmitaiA22,DBLP:conf/aaai/AmitaiSA24}, as well as using LLMs to explain trajectories concerning reward performance~\cite{lu2024mental}, these studies do not address safety.
There exists a variety of related work focusing on the trained RL policy verification~\cite{DBLP:conf/sigcomm/EliyahuKKS21,DBLP:conf/sigcomm/KazakBKS19,DBLP:conf/pldi/ZhuXMJ19,DBLP:conf/seke/JinWZ22}.
Safe RL, such as RL policy shielding~\cite{DBLP:conf/aaai/Carr0JT23}, steers the policy during training to satisfy safety properties.
In comparison, we combine model checking with LLMs to propose explainable alternative actions post-training.
There exists work that combines explainability with safety in explaining NN policy interconnections~\cite{esann2024_gross}.
However, we additionally use the results to improve the safety of the trained policies.

\section{Background}
\paragraph{Probabilistic model checking.} A \textit{probability distribution} over a set $X$ is a function $\mu \colon X \rightarrow [0,1]$ with $\sum_{x \in X} \mu(x) = 1$. The set of all distributions on $X$ is $Distr(X)$.

\begin{definition}[MDP]\label{def:mdp}
A \emph{MDP} is a tuple $M = (S,s_0,Act,Tr, rew,AP,L)$ where
$S$ is a finite, nonempty set of states; $s_0 \in S$ is an initial state; $Act$ is a finite set of actions; $Tr\colon S \times Act \rightarrow Distr(S)$ is a partial probability transition function;
$rew \colon S \times Act \rightarrow \mathbb{R}$ is a reward~function;
$AP$ is a set of atomic propositions;
$L \colon  S \rightarrow 2^{AP}$ is a labeling function.
\end{definition}
We employ a factored state representation where each state $s$ is a vector of features $(f_1, f_2, ...,f_d)$ where each feature $f_i\in \mathbb{Z}$ for $1 \leq i \leq d$ (state dimension).
MDPS can be modeled with the formal language called PRISM\footnote{PRISM manual, http://www.prismmodelchecker.org/manual/}.

A \emph{memoryless deterministic policy $\pi$} for an MDP $M$ is a function $\pi \colon S \rightarrow Act$ that maps a state $s \in S$ to action $a \in Act$.
Applying a policy $\pi$ to an MDP $M$ yields an \emph{induced discrete-time-Markov chain (DTMC)} $D$, where all non-determinism is resolved.
Storm~\cite{DBLP:journals/sttt/HenselJKQV22} allows the verification of PCTL properties of induced DTMCs to make, for instance, safety measurements.
In a slight abuse of notation, we use PCTL state formulas to denote probability values.
For instance, in this paper, $P(\lozenge \text{event})$ denotes the probability of eventually reaching the event.

The standard learning goal for RL is to learn a policy $\pi$ in an MDP such that $\pi$ maximizes the accumulated discounted reward~\cite{DBLP:journals/ml/Bekkemoen24}, that is, $\mathbb{E}[\sum^{N}_{t=0}\gamma^t R_t]$, where $\gamma$ with $0 \leq \gamma \leq 1$ is the discount factor, $R_t$ is the reward at time $t$, and $N$ is the total number of steps.

\paragraph{Large language models.}
In our setting, an LLM is a black-box function that takes input text and outputs text. For details on training LLMs, we refer the reader to~\cite{DBLP:conf/nips/VaswaniSPUJGKP17}.

\section{Methodolodgy}\label{sec:met}
We first incrementally build the induced DTMC of the policy $\pi$ and the MDP $M$ as follows. For every reachable state $s$ via the trained policy $\pi$, we query for an action $a = \pi(s)$. In the underlying MDP $M$, only states $s'$ reachable via that action $a \in A(s)$ are expanded~\cite{DBLP:conf/setta/Gross22}. The resulting DTMC $D$ induced by $M$ and $\pi$ is fully deterministic, with no open action choices, and is passed to the model checker Storm for verification, yielding the \emph{exact} safety measurement result $m$.

Then, we extract all the state-action pairs that led via their transitions to safety violations (based on the PCTL formula).
Afterwards, for each state-action-pair, we input a description of the RL environment, state-action pair, and the question ``What went wrong during execution?'' into an LLM that outputs a human-readable explanation and proposes an alternative action at each stage of the extracted state-action pairs to improve the RL policy performance concerning the safety property (see Example~\ref{ex:input} for LLM input and Example~\ref{ex:output} for the corresponding LLM output).

\begin{example}[LLM input]\label{ex:input}
    \emph{In the Cleaning Agent environment, a robotic agent is tasked with cleaning rooms... Negative state feature values indicates terminal states.
    The action space is: NEXT for changing rooms if room is clean or blocked... CLEAN for cleaning a room...  What went wrong with likelihood {prob} in the state [dirt level=3,blocked=false] with action NEXT ending up in  [dirt level=-1,blocked=false].  Explain it to me.}
\end{example}
\begin{example}[LLM Ouput]\label{ex:output}
    \emph{The agent left the dirty room that nobody was taking care of. An alternative action would be to use the clean action.}
\end{example}
Finally, we rebuild the induced DTMC $D'$ via the alternative actions and verify it again to get a new safety measurement result $m'$.

\paragraph{Limitations.}
This approach focuses on explainable safety repairs fixable one state before a violation and supports memoryless policies in MDP environments, limited by state space and transitions~\cite{DBLP:conf/setta/Gross22}.

\section{Experiments}
We present a private environment (to ensure it was not part of LLM training), trained RL policy, safety repair methods, and technical setup before evaluating explainable safety repairs in various scenarios\footnote{The code is available at \url{https://github.com/LAVA-LAB/COOL-MC/tree/xrl\_llm\_safety}}.

\paragraph{Environment.}
A robotic agent cleans rooms while avoiding collisions and conserving energy. The state includes room cleanliness, slipperiness, and the agent's battery level. The agent is rewarded for correct actions, and the environment terminates upon collisions, energy depletion, or cleaning an already clean room.
\begin{align*}
    S &= \{ \text{(dirt1, dirt2, energy, slippery\text{ }level, room\text{ }blocked)}, \dots \} \\
    Act &= \{ \text{next\text{ }room, charge\text{ }option1, charge\text{ }option2,} \\
    &\quad \text{clean1\text{ }option1, clean1\text{ }option2, clean2\text{ }option1,} \\
    &\quad \text{clean2\text{ }option2, all\text{ }purpose\text{ }clean, idle} \} \\
    \text{rew} &= 
    \begin{cases}
        20 \cdot dirt*, & \text{if clean* operation for dirt* successful.} \\
        20 \cdot dirt1 \cdot dirt2, & \text{if all purpose clean operation successful.} \\
        20, & \text{if changing room correctly.}\\
        10, & \text{if idle when slippery level>0 an room not blocked.}\\
        10, & \text{if charging starts between energy>0 and energy$\leq2$.} \\
        0, & \text{otherwise.}
    \end{cases}
\end{align*}

\paragraph{RL policy training.}
We trained an RL policy using deep Q-learning~\cite{mnih2013playing} with 4 hidden layers of 512 neurons each. Training parameters were a batch size of 64, epsilon decay of 0.99999, minimum epsilon 0.1,  initial epsilon 1, $\gamma$ 0.99, and target network updates every 1024 steps. The policy achieved an average reward of 67.8 over 100 episodes in 27,709 epochs.

\paragraph{Counterfactual safety reasoning methods.}
We compare three counterfactual safety reasoning methods: LLM with a MDP encoded in PRISM, LLM with natural language RL environment description, and a baseline method selecting the second-best policy choice~\cite{DBLP:conf/aaai/AmitaiSA24}. 
We use the state-of-the-art GPT4-turbo API~\cite{DBLP:journals/corr/abs-2303-08774}. We only append the most likely token during text generation to ensure deterministic output and avoid excessive sampling.

\paragraph{Technical setup.}
We executed our benchmarks in a docker container with 16 GB RAM, and an 12th Gen Intel Core i7-12700H $\times$ 20 processors with the operating system Ubuntu 20.04.5 LTS.
For model checking, we use Storm 1.7.0.

\paragraph{Do LLM alternatives enhance safety?}
The results via the different methods are illustrated in Table~\ref{tab:results}.
The LLM, with a natural language explanation of the underlying environment, improved safety performance to avoid running out of energy.
In other cases, our approach matches the baseline's performance but additionally explains why we choose the alternative actions. Inputting the MDP's PRISM encoding into the LLM yields poor results.
One reason could be that LLMs are more trained on natural language than PRISM code.

\begin{table}[h]
\centering
\caption{Different safety policy repair methods and the reachability probability of the unsafe behavior. Lower values are better.}
\label{tab:results}
\begin{tabular}{lrrrr}
\toprule
\textbf{PCTL Query}                      & \textbf{Original} & \textbf{LLM Desc.} & \textbf{LLM PRISM} & \textbf{Baseline} \\ \midrule
P($\lozenge$ no energy)                  & 0.603             & 0.406              & 0.422              & 0.660             \\ 
P($\lozenge$ wrong charge)               & 0.018             & 0.013              & 0.013              & 0.013             \\ 
P($\lozenge$ wrong room switch)          & 0.023             & 0.000              & 0.000              & 0.000             \\ 
P($\lozenge$ wrong idle)                 & 0.023             & 0.000              & 0.050              & 0.000             \\ \bottomrule
\end{tabular}
\end{table}

\paragraph{Are LLM explanations of failures acceptable?}
We analyze the correctness of the action alternatives that were explained. For the safety measure of running out of energy, we reviewed the first 44 explanations. An explanation is deemed correct if both the explanation and proposed action are sensible (we recognize that it may be subjective).
For example, the ratio of correct explanations $P(\lozenge \text{no energy})$ was about 3/4, indicating its potential for explainable policy improvements.

\paragraph{Additional observations}
Action parsing was successful, with only 1 out of 99 LLM outputs having a format issue. 
The LLM approach is highly sensitive to the environment description. For example, with a specific description for P($\lozenge$ no energy), we achieved a probability of 0.265. However, using a more generic description for all PCTL queries (which is used in the Table~\ref{tab:results}), the probability increased to 0.406.

\section{Conclusion}
We used model checking with LLMs for counterfactual safety reasoning, showing LLMs can explain and improve RL policy safety. Future work includes integrating this into safe RL and exploring visual and multi-modal LLMs.

%
%
%
\bibliographystyle{splncs04}
\bibliography{mybibliography}

\end{document}